\PassOptionsToPackage{dvipsnames}{xcolor}
\documentclass[letterpaper, 10 pt, conference]{IEEEtran}
\IEEEoverridecommandlockouts
\usepackage{cite}
\usepackage{amsmath,amssymb,amsfonts}
\usepackage{algorithmic}
\usepackage{graphicx}
\usepackage{textcomp}
\usepackage{xcolor}
\usepackage{subfigure}
\newcounter{MYtempeqncnt}
\usepackage{multirow}
\usepackage{makecell}
\usepackage{color}
\usepackage[dvipsnames]{xcolor}
\usepackage[top=55pt, bottom=55pt, left=54pt, right=54pt]{geometry}

\title{\LARGE \bf
Predicting Driver's Perceived Risk: a Model Based on Semi-Supervised Learning Strategy
}

\author{Siwei Huang, Chenhao Yang and Chuan Hu% <-this % stops a space
\thanks{This work was supported by the Outstanding Youth Science Foundation Project (Overseas) of National Natural Science Foundation of China (NSFC) under grant number 24Z990200855.} 
\thanks{*Chuan Hu is the corresponding author.}% <-this % stops a space
\thanks{Siwei Huang, Chenhao Yang and Chuan Hu are with School of Mechanical Engineering, Shanghai Jiao Tong University, Shanghai, China
        {\tt\small sisi-twt-24@sjtu.edu.cn, chenhaoyan3039@sjtu.edu.cn, chuan.hu@sjtu.edu.cn}}%
}

\begin{document}

\maketitle
\thispagestyle{empty}
\pagestyle{empty}

\begin{abstract}
Drivers' perception of risk determines their acceptance, trust, and use of the Automated Driving Systems (ADSs). However, perceived risk is subjective and difficult to evaluate using existing methods. To address this issue, a driver’s subjective perceived risk (DSPR) model is proposed, regarding perceived risk as a dynamically triggered mechanism with anisotropy and attenuation. 20 participants are recruited for a driver-in-the-loop experiment to report their real-time subjective risk ratings (SRRs) when experiencing various automatic driving scenarios. A convolutional neural network and bidirectional long short-term memory network with temporal pattern attention (CNN-Bi-LSTM-TPA) is embedded into a semi-supervised learning strategy to predict SRRs, aiming to reduce data noise caused by subjective randomness of participants. The results illustrate that DSPR achieves the highest prediction accuracy of 87.91\% in predicting SRRs, compared to three state-of-the-art risk models. The semi-supervised strategy improves accuracy by 20.12\%. Besides, CNN-Bi-LSTM-TPA network presents the highest accuracy among four different LSTM structures. This study offers an effective method for assessing driver’s perceived risk, providing support for the safety enhancement of ADS and driver’s trust improvement.
\end{abstract}

\section{Introduction}\label{s1}
With the development of automated vehicles (AVs), safetyhas been recognized as the key bottleneck limiting the progress and widespread adoption of autonomous driving \cite{wang2024evidence}. Reports of safety problems and accidents continue to hinder public confidence and acceptance of AVs \cite{kenesei2022trust}. How to meet driver's expectations for the safety of automated driving systems (ADSs) becomes a crucial issue \cite{kyriakidis2015public}. Well-considered approaches to perceived risk assessment are needed.

Tradition research mostly focuses on objective risks in traffic scenarios. However, because of ADS's imperfect performance and black-box behavior, driver’s perceived risk may not be consistent with the objective risk in traffic environments. Considering this, risk should be comprehensively assessed from both objective risk and driver’s perceived risk perspectives. The following sections will separately introduce these two perspectives.

\subsection{Objective Risk Assessment and Modeling}\label{s1a}

Crash frequency and severity are important evaluation indicators for the safety of ADS \cite{wang2021review}. However, collisions are rare events. Thus, specific and quantifiable criteria are applied to evaluate objective risk. Common methods are rule-based methods and learning-based methods \cite{song2024subjective}.

Rule-based methods assess risk using objective indicators from the traffic environment, with Surrogate Safety Measures (SSMs) and Risk Field (RF) being two common approaches. SSMs describe the conflicts between users in the traffic environment in terms of time and space, such as time to collision (TTC) and time headway (TWH). RF generally considers the kinematics of all traffic objects to build risk models. Wang et al. \cite{wang2015driving} developed a Driving Safety Field (DSF) model by overlaying potential fields, kinetic fields, and behavior fields. Chen et al. \cite{chen2024quantifying} proposed a Potential Damage Risk (PODAR) describing potential damage by energy and decay functions.

However, rule-based methods perform poorly in assessing coupled risks of complex traffic environment. Machine learning methods have a distinct advantage under this condition. Xie et al. \cite{xie2024real} collected driver’s operation, visual, physiological, and vehicle kinematic data, predicting risk levels through an Att-Bi-LSTM network. Liu et al. \cite{liu2023learning} used a Transformer network to capture complex interaction dynamics between pedestrians and vehicles to predict their collision risks.

\subsection{Driver’s Perceived Risk Assessment and Modeling}\label{s1b}
Drivers form judgements about the risk when they are in traffic environment. Limited attention and working memory hinder them from acquiring and interpreting all information\cite{endsley1995toward}. That’s why driver's subjective perceived risk often does not align with the objective risk in traffic environment. 

Investigating driver’s perceived risk requires collecting feedback from the human driver, where driver-in-loop experiments are essential. Simulator experiments \cite{he2022modelling}, real vehicle experiments \cite{stapel2022road}, and video experiments \cite{ping2022enhanced} are conducted to provide driving scenarios. Interval rating \cite{yu2021scene} and continuous rating through equipment \cite{ping2022enhanced} are conducted to obtain driver's real-time subjective risk ratings (SRRs). For example, Ping et al. \cite{ping2022enhanced} presented real-world driving videos to drivers while continuously collecting their SRRs via computer buttons. They extracted a semantic understanding of traffic scenes from visual images and used a GRU network to predict driver's perceived risk. Yu et al. \cite{yu2021scene} collected SRRs ranging from -2 to 2 for lane-change segments through simulator experiment. They used a graph convolution network and LSTM network to predict driver's perceived risk.

Objective indicators from traffic environment are complex and unpredictable. Some studies suggest preliminary modeling of these indicators before integrating them with drivers' SRRs for further analysis. For example, based on RF model and spatiotemporal distribution of driver's cognitive risk, Song et al. \cite{song2024subjective} developed a machine learning-based method for predicting subjective risk in lane-changing scenarios.

To reduce noise and improve consistency of SRRs, some studies have also introduced semi-supervised learning \cite{zhu2022comfort}. Additionally, semi-supervised learning is effective in learning risk patterns from a small amount of labeled data combined with a large volume of unlabeled data \cite{hu2021cost}. However, few studies have applied it to driver’s perceived risk assessment.

\subsection{The Current Research}\label{s1c}
Although existing research has made significant progress in objective risk assessment, there is still limited research on driver’s perceived risk. Furthermore, few studies have integrated objective and subjective risks for a comprehensive analysis. To address these issues, we propose a driver’s perceived risk (DSPR) model and utilize machine learning methods to predict driver's SRRs. The main contributions of this paper are as follows: (1) Proposed DSPR model considers both objective risk and driver's risk perception patterns, with a risk-trigger mechanism introduced; (2) We introduce a semi-supervisor learning strategy with a convolutional neural network and bidirectional long short-term memory network with temporal pattern attention (CNN-Bi-LSTM-TPA) for driver's SRRs prediction. 

\begin{figure}[t]
\centerline{\includegraphics[width=0.90\linewidth]{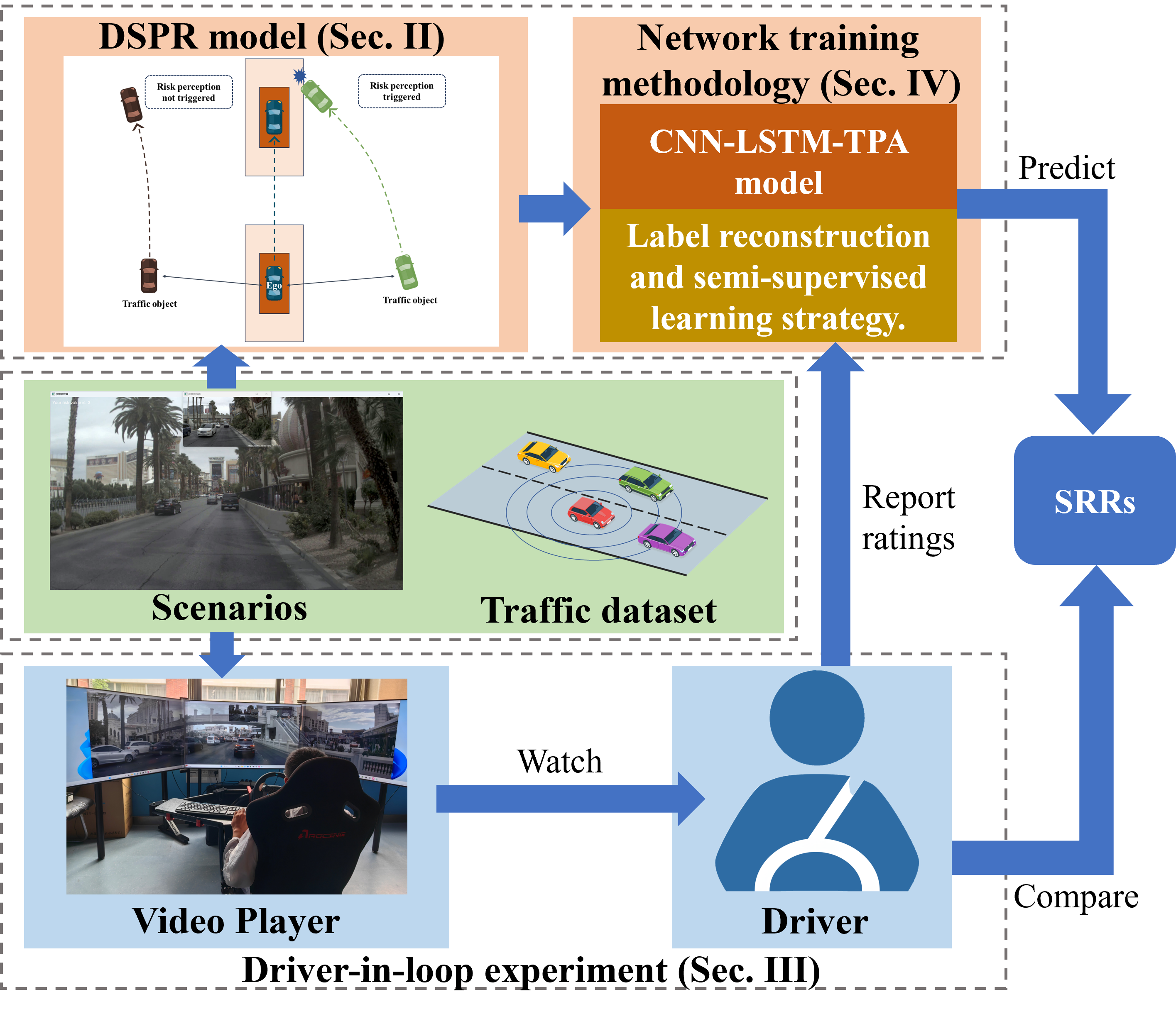}}
\vspace{-0.45em}
\caption{Framework of our study.}
\label{fig1}
\vspace{-1em}
\end{figure}

The rest of this paper is organized as follows: Section \ref{s2} introduces the established DSPR model. Section \ref{s3} introduces the experiment and the data collection method. Section \ref{s4} introduces the network training methodology. Section \ref{s5} presents results and discussion. Section \ref{s6} concludes this work. The framework of our study is shown in Fig.~\ref{fig1}.

\section{The Modelling of Driver's Perceived Risk}\label{s2}

According to the situation awareness (SA) theory \cite{endsley1995toward}, driver's perception of the situation occurs at three levels: 1) Perception of the elements in the environment; 2) Comprehension of the current situation; 3) Projection of future status. Most existing RF models describe the second level, while neglecting the other two. Considering this issue, we propose DSPR to considers both objective risk and driver’s perception patterns. Incorporating the concept from \cite{chen2024quantifying}, we define driver’s perceived risk as “the expectation and assessment of the likelihood and severity of a risk event.” We introduce the risk-trigger mechanism for the first level, observation sensitivity for the second level, and time and spatial decay of risk for the third level of SA theory. The basic formula of DSPR is shown in \eqref{eq1}:
\begin{equation}
\textbf{R}(t) = \begin{bmatrix} 
\mu_1 R_1(t) & \dots & \mu_i R_i(t) & \dots & \mu_n R_n(t) 
\end{bmatrix}\label{eq1}
\end{equation}
where \(\textbf{R}(t)\) is the perceived risk sequence calculated from DSPR; \(R_i (t)\) is the risk of traffic object \(O_i\) at current timestamp \(t\); \(\mu_i\) is the sensitivity coefficient of \(O_i\) ; \(n\) is the total number of considered traffic objects.

\subsection{Risk-trigger Mechanism}\label{s2a}
\begin{figure*}[t]
\centering
\begin{minipage}{0.36\linewidth}
    \centering
    \includegraphics[height=5.2cm]{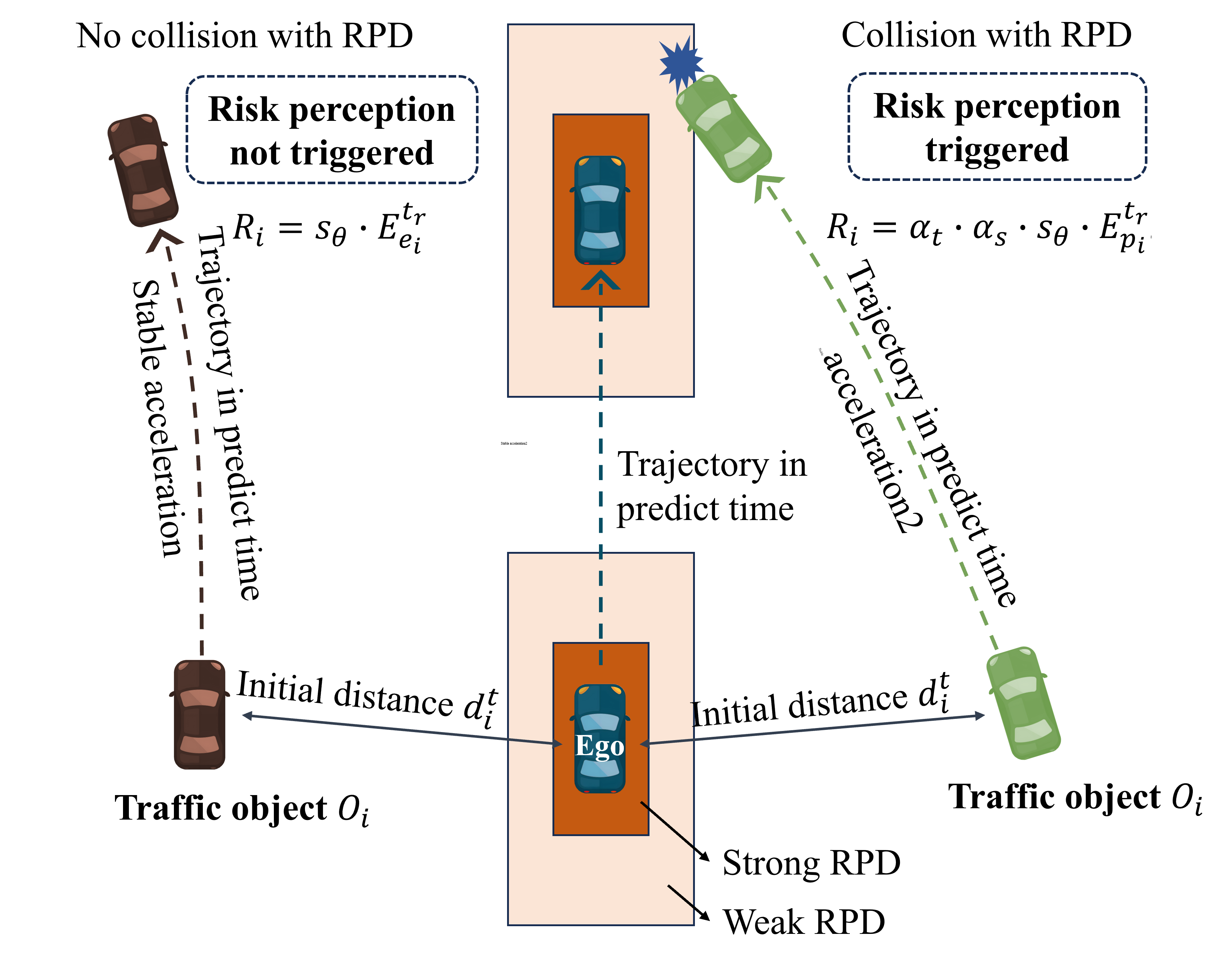}
    \caption{The RPD and the risk perception triggering mechanism.}
    \label{fig2}
\end{minipage}%
\hspace{0.4cm} % 控制图像间的水平间隙
\begin{minipage}{0.32\linewidth}
    \centering
    \includegraphics[height=5cm]{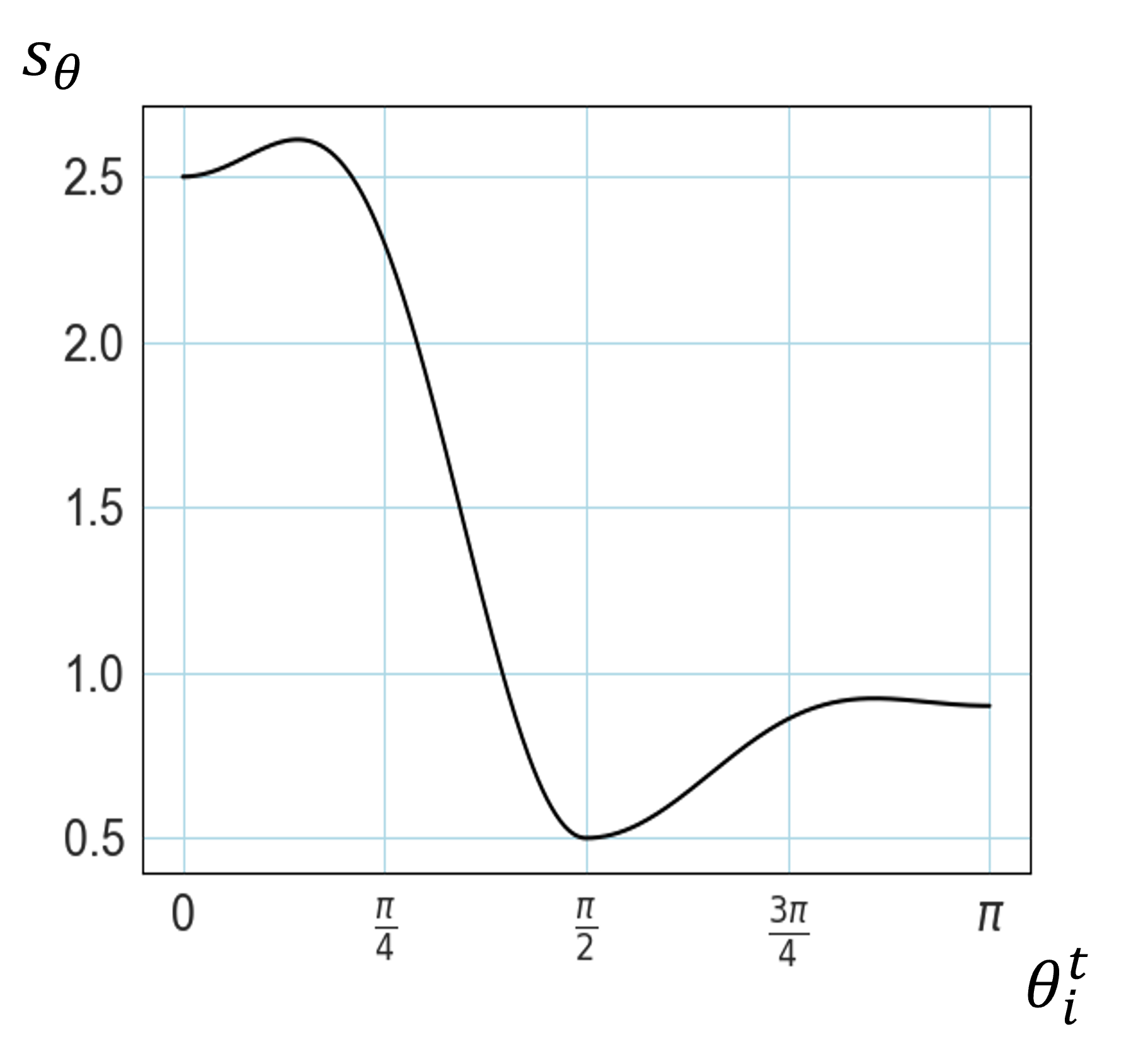}
    \caption{The relationship between driver’s observation sensitivity \(s_\theta\) and initial angle \(\theta_i^t\) of traffic object \(O_i\).}
    \label{fig4}
\end{minipage}%
\hspace{0.4cm} % 控制图像间的水平间隙
\begin{minipage}{0.24\linewidth}
    \centering
    \includegraphics[height=5.08cm]{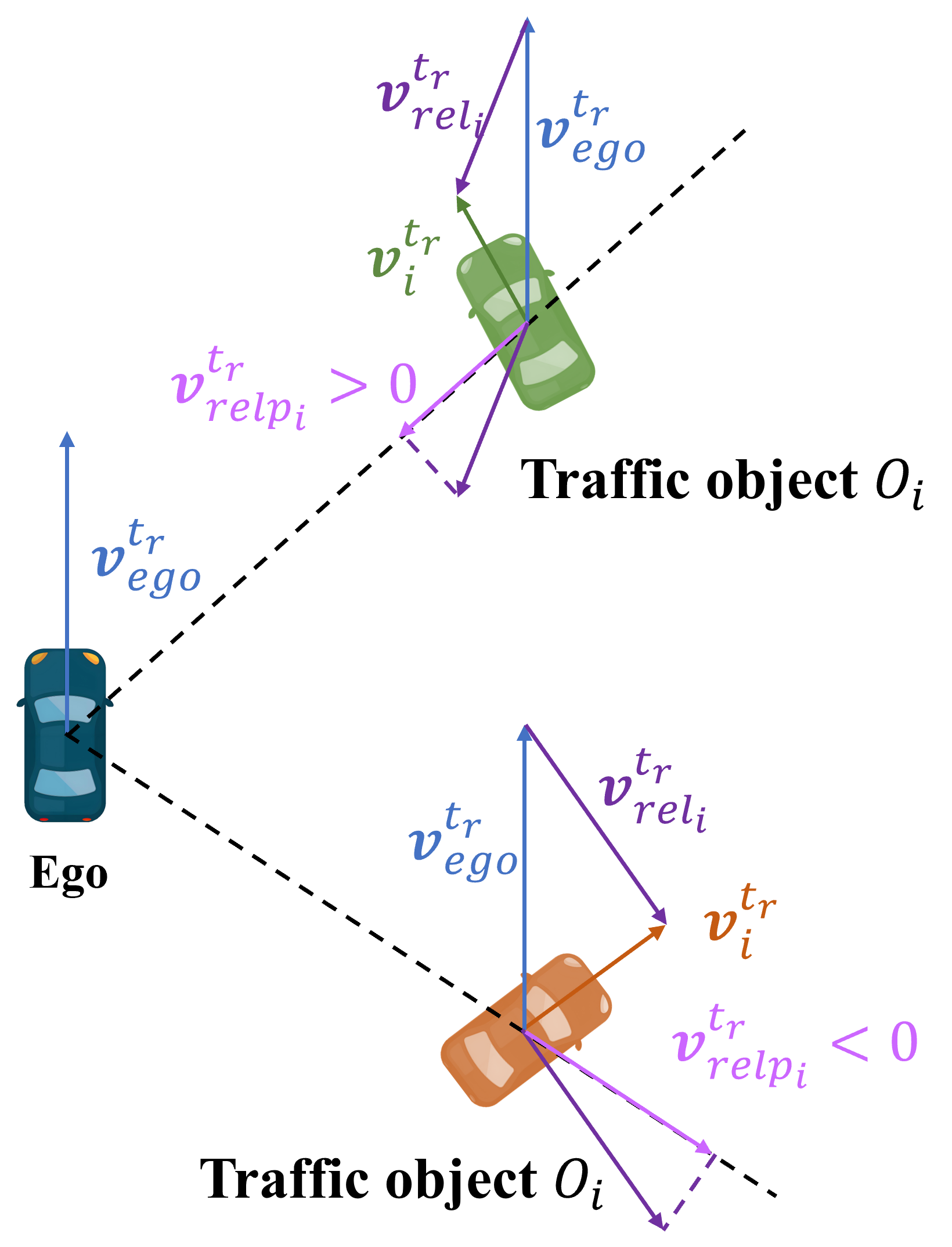}
    \caption{The calculation methods for \(v_{\text{rel}_i}^{t_r}\) and \(v_{\text{relp}_i}^{t_r}\).}
    \label{fig5}
\end{minipage}
\vspace{-1em}
\end{figure*}

The concept of the risk-trigger mechanism stems from the idea that a driver’s attention and perception are limited. As shown in Fig.~\ref{fig2}, we assume that only traffic object \(O_i\) entering a dynamic risk perception domain (RPD) within the prediction time \(t_p\) from current moment will cause the driver to perceive risk (risk perception is triggered). It can prevented risk from extending indefinitely, focusing on the risks that related to the ego vehicle. We assume that the acceleration of the ego vehicle and traffic objects remain stable within \(t_p\).

If the risk perception is triggered, \(R_i\) is defined in \eqref{eq2}.
\begin{equation}
R_i(t) = \alpha_t \cdot \alpha_s \cdot s_\theta \cdot E_{p_i}^{t_r}\label{eq2}
\end{equation}
where \(\alpha_t\) is the time decay coefficient; \(\alpha_s\) is the spatial decay coefficient; \(s_\theta\) is the driver’s observation sensitivity; \(E_{p_i}^{t_r}\) is the relative kinetic energy. 

If the risk perception is not triggered, the driver will only perceive risk from the ego vehicle. \(R_i\) is defined in \eqref{eq3}.
\begin{equation}
R_i(t) = s_\theta \cdot E_{e_i}^{t_r}\label{eq3}
\end{equation}
where \(E_{e_i}^{t_r}\) is the relative kinetic energy of ego vehicle.

%\begin{figure*}[t]
%\centering
%\begin{minipage}{0.48\linewidth}
%    \centering
%    \includegraphics[height=5.6cm]{Fig2-DSPR.png}
%    \caption{The RPD and the risk perception triggering mechanism.}
%    \label{fig2}
%\end{minipage}%
%\begin{minipage}{0.48\linewidth}
%    \centering
%    \includegraphics[height=5.6cm]{Fig3-angle.png} 
%    \caption{The angle of the collision position \(\theta_{s_i}\) (not used) and \(\theta_{w_i}\), and the initial angle \(\theta_i^t\) of traffic object \(O_i\).}
%    \label{fig3}
%\end{minipage}%
%\vspace{-1em}
%\end{figure*}

%\begin{figure}[t]
%\centerline{\includegraphics[width=1\linewidth]{Fig2-DSPR.png}}
%\vspace{-0.5em}
%\caption{The RPD and the risk perception triggering mechanism.}
%\label{fig2}
%\vspace{-1.5em}
%\end{figure}

To simplify calculation, we consider the shape of vehicles and RPDs as a rectangle. The shape of the RPD is variable. The idea comes from THW. For longitudinal risk, driver's perception distance increase with ego vehicle’s velocity, while almost doesn’t change for lateral risk. We define the weak RPD (w-RPD) where driver perceives risk, and the strong RPD (s-RPD) where driver strongly perceives risk. We denote the length of ego vehicle as \(L\), width as \(W\), velocity as \(v_{ego}\) , the time headway threshold of the w-RPD and s-RPD as \(\text{THW}_w\) and  \(\text{THW}_s\). The length of the w-RPD is \(l_w = 2 \left( L + \text{THW}_w \cdot v_{\text{ego}} \right)\); the width of the w-RPD is \(w_w\ = 5W\); the length of s-RPD is \(l_s = 2 \left( L + \text{THW}_s \cdot v_{\text{ego}} \right)\); the width of s-RPD is \(w_s\ = 2W\).

\subsection{Time Decay and Spatial Decay of Risk}\label{s2b}
\textit{1) Time decay coefficient.} We denote the time from the current timestamp \(t\) to the time of risk activation as \(t_r\). The earlier this moment occurs, the more clearly the driver perceives the risk. When \(O_i\) enters RPD at \(t_r\) within \(t_p\), the time decay coefficient \(\alpha_t\) is defined in \eqref{eq4}.
\begin{equation}
\alpha_t = \frac{t_p}{t_p + t_r}
\label{eq4}
\end{equation}

\textit{2) Spatial decay coefficient.} The spatial decay coefficient \(\alpha_{s}\) reflects how the distance from a traffic object affects the driver's perception of risk. As shown in \eqref{eq5},  \(\alpha_{ss}\) is the coefficient of s-RPD, \(\alpha_{ss}\) is the coefficient of w-RPD \(\alpha_{ws}\). 
\begin{equation}
\alpha_s=\alpha_{ss}\cdot\alpha_{ws}
\label{eq5}
\end{equation}

Drivers have different tolerances for safe distances in various directions. We use a sine function to model the differences in tolerance across directions. \(d_i^t\) is the initial distance between ego vehicle and \(O_i\) at \(t\). The straight-ahead of ego vehicle is \(0^\circ\), and rear is \(180^\circ\). \(\theta_i^t\) is the initial angle of \(O_i\) from from the ego vehicle’s perspective, \(\theta_{w_i}\) is the angle of the collision position with the w-RPD (if collision happens). Within \(t_p\), if \(O_i\) collides with the s-RPD, We define that \(\alpha_{ss} = \frac{l_{\text{strong}} - \sin \theta_i^t \cdot (l_{\text{strong}} - w_{\text{strong}})}{d_i^t}\) and \(\alpha_{ws}=1\). If \(O_i\) collides with the w-RPD, \(\alpha_{ss}\) keeps the same and \(\alpha_{ws} = \frac{l_{\text{strong}} - \sin \theta_{w_i} \cdot (l_{\text{strong}} - w_{\text{strong}})}{d_i^{t_r}}\).

\subsection{Driver’s observation sensitivity.}\label{s2c}
Driver’s observation exhibits anisotropy. Their attention usually focuses on the road ahead, leading to neglect of other areas. The observation sensitivity \(s_\theta\) can balance the weight of risk among surrounding traffic objects in \eqref{eq6}.
As shown in Fig.~\ref{fig4}, for \(0^\circ \) to \(30^\circ\), traffic objects are easily noticeable, so \(s_\theta\) is at its maximum. For \(30^\circ \) to \(75^\circ\), traffic objects are gradually not easy to notice. For \(75^\circ \) to \(150^\circ\), the driver needs to turn their head to observe around, and \(s_\theta\) gradually reaches its minimum value. For \(150^\circ \) to \(180^\circ\), driver can observe through the rearview mirror, and \(s_\theta\) is above the minimum value. 

%\begin{figure*}[t]
%\centering
%\begin{minipage}{0.48\linewidth}
%    \centering
%    \includegraphics[height=5.6cm]{Fig4-s-theta.png}
%    \caption{The relationship between driver’s observation sensitivity \(s_\theta\) and initial angle \(\theta_i^t\) of traffic object \(O_i\).}
%    \label{fig4}
%\end{minipage}%
%\begin{minipage}{0.48\linewidth}
%    \centering
%    \includegraphics[height=5.6cm]{Fig5-v.png}
%    \caption{The calculation methods for \(v_{\text{rel}_i}^{t_r}\) and \(v_{\text{relp}_i}^{t_r}\).}
%    \label{fig5}
%\end{minipage}
%\vspace{-0.5em}
%\end{figure*}

\begin{figure*}[!t] % hb底部，ht为头部
\normalsize
% Store the current equation number.
\setcounter{MYtempeqncnt}{\value{equation}}
% Set the equation number to one less than the one
% desired for the first equation here.
% The value here will have to changed if equations
% are added or removed prior to the place these
% equations are referenced in the main text.
\setcounter{equation}{5}
\vspace{+0.5em}
\begin{equation}
s_{\theta} = 
\begin{cases} 
A_{\theta} \left( \cos(2\theta_i^t) + 1 \right) + B_{\theta} \left( 1 - \cos(4\theta_i^t) \right) + C_{\theta}, & 0^\circ < \theta < 90^\circ \\
A_{\theta} \left( \cos(2\theta_i^t - \pi) + 1 \right) + B_{\theta} \left( 1 - \cos(4\theta_i^t - \pi) \right) + C_{\theta}, & 90^\circ < \theta < 180^\circ
\end{cases}
\label{eq6}
\end{equation}
\vspace{-0.1em}
\begin{equation}
E_{p_i}^{t_r} = 
\begin{cases} 
\frac{1}{2} m_{s_i} \left( (1-\beta) v_{\text{relp}_i}^{t_r} + \beta v_{b_i}^{t_r} \right) \cdot \left( \beta \left( v_{\text{rel}_i}^{t_r} + v_{b_i}^{t_r} \right) \right), & x < 0 \\
\frac{1}{2} m_{s_i} \left( \beta v_{b_i}^{t_r} \right) \cdot \left( \beta \left( v_{\text{rel}_i}^{t_r} + v_{b_i}^{t_r} \right) \right), & x \geq 0 
\end{cases}
\label{eq7}
\end{equation}
\vspace{-0.5em}
\setcounter{equation}{\value{MYtempeqncnt}}
% The IEEE uses as a separator
\hrulefill
\vspace{0em}
\end{figure*}

\subsection{Relative Kinetic Energy.}\label{s2d}
We use a concept of kinetic energy to describe the driver’s expectations regarding collision consequences. The relative kinetic energy \(E_{p_i}^{t_r}\) is defined in \eqref{eq7}. 
Where \(v_{b_i}^{t_r} = v_{\text{ego}}^{t_r} + v_i^{t_r}\), \(m_{s_i}\) is the coefficient regarding the type of \(O_i\); \(\beta\) is the balancing coefficient of velocity and relative velocity. \(v_{\text{relp}_i}^{t_r}\) is the projection of the relative velocity between the ego vehicle and \(O_i\) in the direction of their relative position; \(v_{\text{rel}_i}^{t_r}\) is the magnitude of the relative velocity vector difference between the ego vehicle and \(O_i\). \(v_{\text{rel}_p}^{t_r}\) and \(v_{\text{relp}_i}^{t_r}\) are shown in Fig.~\ref{fig5}. 

On the other hand, if traffic object \(O_i\) does not collide with any RPD (the risk perception is not triggered), only the background energy \(E_{p_i}^{t_r}\) influenced by the ego vehicle's energy will exist. \(E_{p_i}^{t_r}\) is defined in \eqref{eq8}.
\setcounter{equation}{7}
\begin{equation}
E_{e_i}^t=1/2 m_{s_i} (\beta v_{ego}^{t_r} )^2
\label{eq8}
\end{equation}

\section{Experiment and Data Collection}\label{s3}
\subsection{Data Preparation}\label{s3a}
\begin{figure}[t]
\centerline{\includegraphics[width=0.93\linewidth]{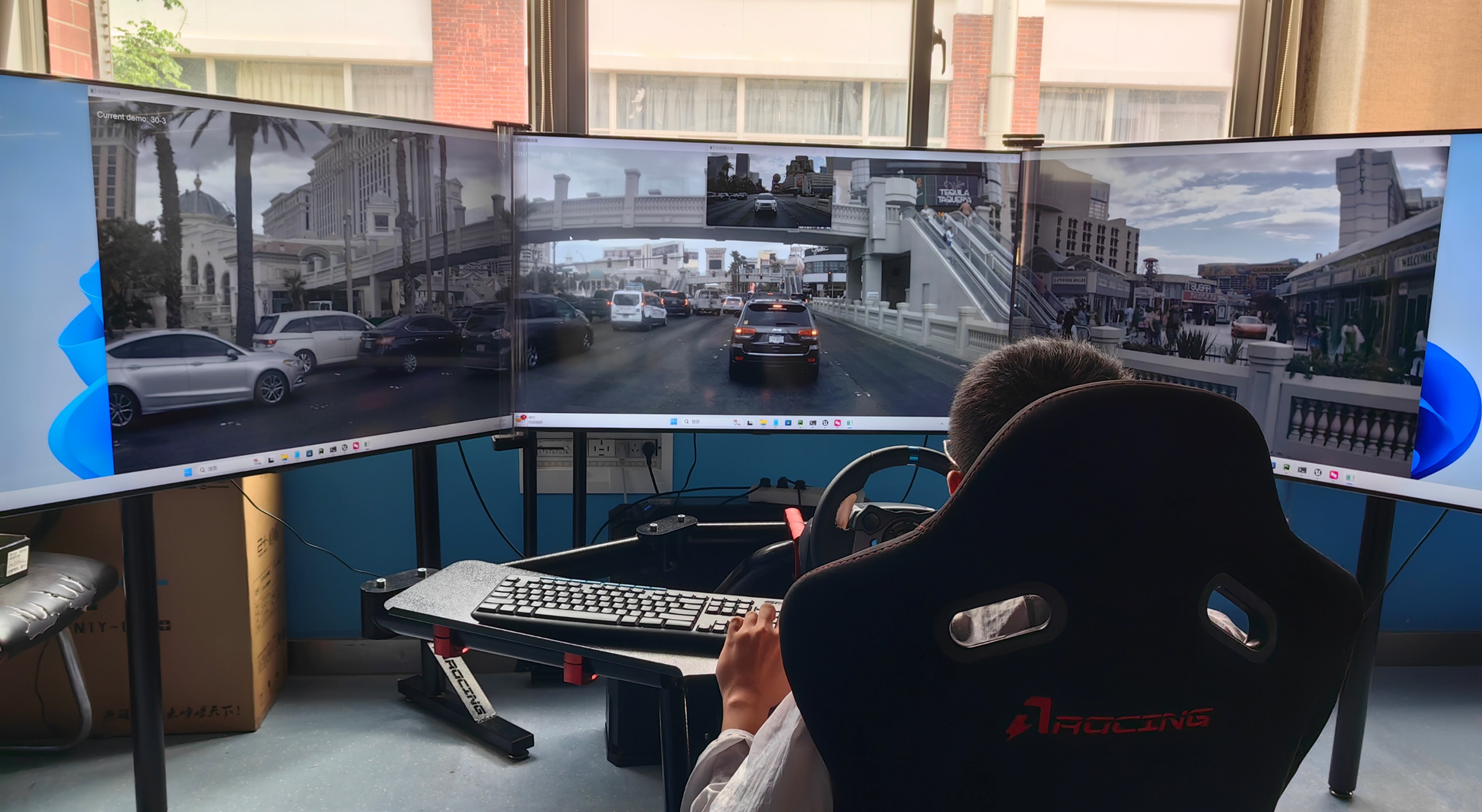}}
\vspace{-0.4em}
\caption{Interface for video playback and rating during the experiment. The platform is based on a triple-screen simulator.}
\label{fig6}
\vspace{-0.4em}
\end{figure}

We chose the NuPlan public dataset \cite{caesar2021nuplan} as the scenario. The dataset includes extensive labeled position and kinematic information of the ego vehicle and surrounding traffic objects, and 360-degree camera view of the ego vehicle. The frequency of the video is 10 Hz. We extracted 113 video clips, with a total duration of 1 hour and 14 minutes. As shown in Fig~\ref{fig6}, we extracted the forward, left-front, right-front, and the mirror-rearward views and stitched them together to play on a three-screen simulator. We did not provide other views because they are blind spots for drivers. This can make participants obtain information closely resembles the actual driving process.

\subsection{Experiment Procedure}\label{s3b}
We recruited 20 participants for SRR collection (5 females and 15 males). All participants hold driving licenses in China. The average driving age of participants is 4.45, and SD is 2.87.

Before the experiment, each driver received an adaptation training to familiarize the rating criteria. During the experiment, real-time SRRs are recorded through the keyboard. The video only plays when the driver holds down a specific rating key, where SRR-1 represents no risk, SRR-2 almost no risk, SRR-3 slight risk, SRR-4 risk, and SRR-5 high risk \cite{ping2022enhanced}. We encouraged drivers to provide ratings as quickly as possible based on their intuition. Besides, we asked them to switch ratings quickly to ensure continuous SRRs.

\subsection{Data Pre-processing}\label{s3c}

As shown in Table~\ref{tab1}, the original data includes the kinematic data of ego vehicle and surrounding traffic objects. For timestamp \(t\), we extract the data of up to 30 closest vehicles and 10 closest pedestrians and input them into DSPR model to obtain the risk matrix \(\textbf{R}(t)\). The size of \(\textbf{R}(t)\) is \(1 \times 40\). Road conditions also affect driver's risk perception. We classify them by one-hot encoding: 1-three or fewer lane road, 2-four or more lane road, 3-pavement, 4-controlled intersection, 5-uncontrolled intersection, and 6-mixed-traffic road.The road condition parameter is denote as \(C\).

\begin{table}[t]
\caption{Original data extracted from Nuplan dataset}
\vspace{-1.4em}
\begin{center}
\begin{tabular}{cc|cc}
\hline
\multicolumn{2}{c|}{\textbf{Ego vehicle data}}&\multicolumn{2}{|c}{\textbf{Traffic data}} \\
\hline
\textbf{\textit{Parameter name}} & \textbf{\textit{Symbol}}& \textbf{\textit{Parameter name}}& \textbf{\textit{Symbol}} \\
\hline
Position & \(x_x\), \(x_y\) & Position of \(O_i\) & \(x_{x_i}\), \(x_{y_i}\) \\
Velocity & \(v_x\), \(v_y\) & Velocity of \(O_i\) & \(v_{x_i}\), \(v_{y_i}\) \\
Acceleration & \(a_x\), \(a_y\) & Acceleration  of \(O_i\) & \(a_{x_i}\), \(a_{y_i}\) \\
Angle & \(\varphi\) & Angle of \(O_i\) & \(\varphi_i\) \\
- & - & Type of \(O_i\) & \(\text{type}_i\) \\
\hline
\end{tabular}
\label{tab1}
\end{center}
\vspace{-1.8em}
\end{table}

The parameters of the ego vehicle, road conditions, and the risk matrix \(\textbf{R}(t)\) are combined to another matrix, denoted as \(\boldsymbol{\Phi}_t\) and defined  in \eqref{eq9}. The size of \(\boldsymbol{\Phi}_t\) is \(1 \times 46\). We combine \(\boldsymbol{\Phi}_t\) of every \(T\) timesteps and denote the network input sequences as \(\boldsymbol{\Psi}_t\), as shown in \eqref{eq10}. Then, we put \(\boldsymbol{\Psi}_t\) into the network. 
\begin{equation}
\boldsymbol{\Phi}_t = \begin{bmatrix} 
v_x & v_y & a_x & a_y & \varphi & C & \textbf{R}(t)
\end{bmatrix}
\label{eq9}
\end{equation}
\begin{equation}
\boldsymbol{\Psi}_t = \begin{bmatrix} 
\boldsymbol{\Phi}_{t-T+1} & \boldsymbol{\Phi}_{t-T+2} & \dots & \boldsymbol{\Phi}_{t-1} & \boldsymbol{\Phi}_t 
\end{bmatrix}^T
\label{eq10}
\end{equation}

\section{Network Training Methodology}\label{s4}
\subsection{CNN-Bi-LSTM-TPA Network for SRR Prediction}\label{s41}
\begin{figure}[t]
\centerline{\includegraphics[width=0.87\linewidth]{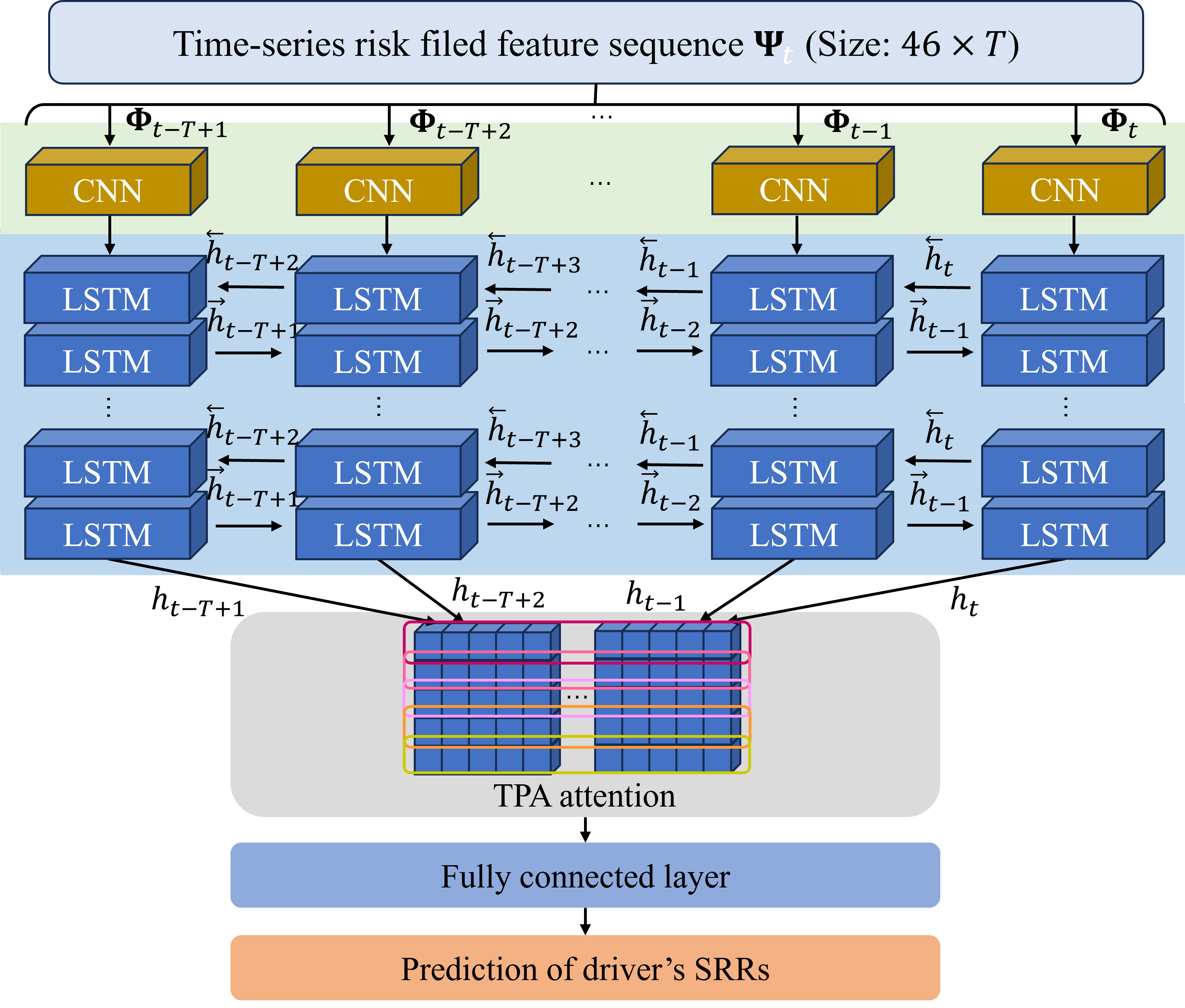}}
\vspace{-1em}
\caption{Structure of CNN-Bi-LSTM-TPA network.}
\label{fig7}
\vspace{-1em}
\end{figure}

Driver’s attention is limited and cannot notice the risk posed by all traffic objects. Besides, it is necessary to consider the time-dependent relationship of risks. Long short-term memory (LSTM) networks can effectively address this issue \cite{hochreiter1997long}. LSTM is widely used in time series processing, and can better capture long-term dependencies in sequences. In our study, we proposed the CNN-Bi-LSTM-TPA network to predict driver’s SRRs. The structure is shown in Fig.~\ref{fig7}.

\textit{1) CNN layer}. Every CNN receives data from a single time step in \(\boldsymbol{\Psi}_t\). It consists of a convolutional layer, a ReLU activation function layer, and a dropout layer. CNN layer helps the network better understand hidden relationships between features.

\textit{2) Bi-LSTM layer}. The Bi-LSTM layer processes the output of CNN layer. Here, \(\overset{\rightarrow}{h}_t\) and \(\overset{\leftarrow}{h}_t\) represent the forward and backward hidden states, respectively. After passing through several stacked Bi-LSTM layers, the hidden state \(h_t\) of the final layer is output to the attention mechanism.

\textit{3) TPA layer}. We combined the attention mechanisms proposed by Shih et al. \cite{shih2019temporal} The TPA mechanism places greater emphasis on the correlation of features in the temporal domain than traditional attention mechanism.

\textit{4) Fully connected (FC) layer}. The output from the attention layer is processed through an FC layer, which ultimately yields the predicted SRRs.

\subsection{Label Reconstruction and Semi-supervised Learning}\label{s4b}
Not all SRRs are credible because of driver’s subjectivity. To improve the consistency and decrease the noise caused by individual variances of SRRs, we define the consistency ratio \(p_T^t\) in \eqref{eq11}. Only if \(p_T^t\) reaches a certain threshold \(p_{\text{true}}\), the rating is marked as a true label.
\begin{equation}
p_T^t = \max \left( \frac{N_r^t}{N} \right) \quad (r = 1, 2, 3, 4)
\label{eq11}
\end{equation}
where \(N_r^t\) is the number of participants of the rating \(r\) at \(t\). \(N\) is the total number of participants. 

According to split ratio \(s\), data are divided into the training set \(\boldsymbol{R}^{\text{train}}\) and the testing set \(\boldsymbol{R}^{\text{test}}\). They are further divided into the true set and the unlabeled set: \(\boldsymbol{R}^{\text{train}} = \{ \boldsymbol{R}_T^{\text{train}}, \boldsymbol{R}_U^{\text{train}} \}\) and \(\boldsymbol{R}^{\text{test}} = \{ \boldsymbol{R}_T^{\text{test}}, \boldsymbol{R}_U^{\text{test}} \}\) by \eqref{eq11}. At \(t\), if \(p_T^t \geq p_{\text{true}}\), it will be included in the true set. Otherwise, included in the unlabeled set. Input sequences and labels are denoted as \(\boldsymbol{D}\) and \(\boldsymbol{L}\): \(\boldsymbol{R}_T^{\text{train}} = \{ \boldsymbol{D}_T^{\text{train}}, \boldsymbol{L}_T^{\text{train}} \}\), \(\boldsymbol{R}_T^{\text{test}} = \{ \boldsymbol{D}_T^{\text{test}}, \boldsymbol{L}_T^{\text{test}} \}\), and \(\boldsymbol{R}_U^{\text{test}} = \{ \boldsymbol{D}_U^{\text{test}}\}\).

\begin{figure}[t]
\centerline{\includegraphics[width=1\linewidth]{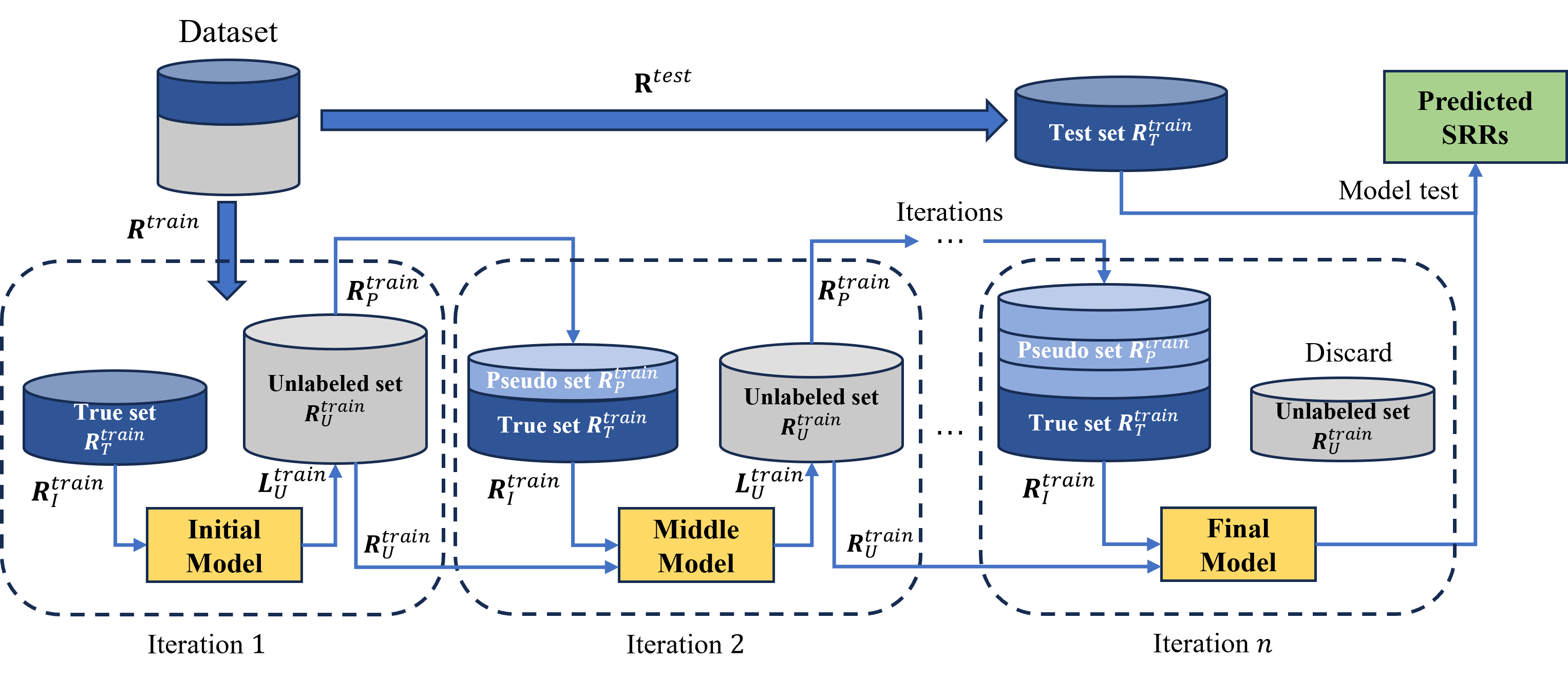}}
\vspace{-1.0em}
\caption{Semi-supervised learning strategy.}
\label{fig8}
\vspace{-0.1em}
\end{figure}

Semi-supervised learning are applied for label reconstruction and updating, as shown in Fig.~\ref{fig8} The steps are as follows: 1) Define the input set as \(\boldsymbol{R}_I^{\text{train}} = \{ \boldsymbol{R}_T^{\text{train}}\}\) to train an initial model. 2) Input \(\boldsymbol{D}_U^{\text{train}}\) into the model to generate new label set \(\boldsymbol{L}_U^{\text{train}}\). 3) Labels in \(\boldsymbol{L}_U^{\text{train}}\) with confidence higher than \(\epsilon\) are selected, denoted as \(\boldsymbol{L}_P^{\text{train}}\). \(\boldsymbol{L}_P^{\text{train}}\) and their corresponding input sequences \(\boldsymbol{D}_P^{\text{train}}\) are added into the pseudo-labeled set \(\boldsymbol{R}_P^{\text{train}} = \{ \boldsymbol{D}_P^{\text{train}}, \boldsymbol{L}_P^{\text{train}} \}\). Labels with confidence lower than \(\epsilon\) are returned to \(\boldsymbol{R}_U^{\text{train}}\). 4) Define \(\boldsymbol{R}_I^{\text{train}} = \{ \boldsymbol{R}_T^{\text{train}}, \boldsymbol{R}_P^{\text{train}} \}\) to train a middle model and repeat step 2) to 4) for several times. 5) After several iterations, discard the rest unlabeled set \(\boldsymbol{R}_U^{\text{train}}\). 6) Define \(\boldsymbol{R}_I^{\text{train}} = \{ \boldsymbol{R}_T^{\text{train}}, \boldsymbol{R}_P^{\text{train}} \}\) to train a final model. 7) Use \(\boldsymbol{R}_T^{\text{test}}\) to test the final model.

\section{Results And Discussion}\label{s5}
\subsection{Parameter}\label{s5a}
\textit{1) DSPR parameters.} Overall, due to the presence of the LSTM network and the attention mechanism, we don't need to focus too much on the absolute value of DSPR. Instead, the relative risk between different situations is what matters most. DSPR parameters are shown in Table~\ref{tab3}.

\begin{table}[t]
\caption{Parameters of DSPR.}
\vspace{-2.25em}
\begin{center}
\resizebox{\linewidth}{!}{
\begin{tabular}{cccccccc}
\hline
\textbf{Para} & \textbf{Value}& \textbf{Para} & \textbf{Value} & \textbf{Para} & \textbf{Value} & \textbf{Para} & \textbf{Value} \\
\hline
\(L\) & \(4.8\) & \(A_\theta\) & \(1\) & \(\beta\) & \(0.12\) & \(\text{THW}_w\) & \(2.4\) \\
\(W\) & \(2\) & \(B_\theta\) & \(0.4\) & \(m_{s_i}\)(v) & \(5\) & \(\text{THW}_s\) & \(1.2\) \\
\(t_p\) & \(4\) & \(C_\theta\) & \(0.5\) & \(m_{s_i}\)(p) & \(10\) & - & - \\
\hline
\multicolumn{8}{l}{\(m_{s_i}\)(v) is \(m_{s_i}\) for vehicle. \(m_{s_i}\)(p) is \(m_{s_i}\) for pedestrian.}
\end{tabular}
}
\label{tab3}
\end{center}
\vspace{-1.5em}
\end{table}

\textit{2) Training parameters.}  We set \(T\) to 10 (equal to 1 second). split ratio for train set \(s=0.7\). Considering the imbalanced amount of different SRRs, we apply different \(p_{true}\) values to each SRR and the SMOTE-NC (Synthetic Minority Over-sampling Technique for Nominal and Continuous) method to augment the minority classes. The \(p_{true}\) values and statistics of SRR labels are shown in Table~\ref{tab5}. 
Because participants provided significantly fewer ratings for SRR-4 and SRR-5 (Just 2\% among all SRR labels). Distinguishing between SRR-4 and SRR-5 may even result in certain type of SRRs being absent from the training or test sets. Therefore, we merged SRR-4 and SRR-5 into a single level, represented by SRR-4 as indicating risk. After processed by SMOTE-NC, each SRR category is adjusted to the same number of label instances (477 true labels for train set and 184 true labels for test set in each SRR category).

\begin{table}[t]
\caption{The \(p_{true}\) values and statistics of SRRs labels.}
\vspace{-2em}
\begin{center}
\begin{tabular}{ccccc}
\hline
\textbf{Dataset} & \textbf{SRR-1} & \textbf{SRR-2} & \textbf{SRR-3} & \textbf{SRR-4*} \\
\hline
\textbf{Total} & 1403 & 2013 & 915 & 115\\
\(\boldsymbol{p_{\text{true}}}\) & 0.9 & 0.75 & 0.65 & 0.6\\
\textbf{True Label in Train Set} & 477 & 35 & 36 & 9 \\
\textbf{True Label in Test Set.} & 184 & 17 & 11 & 6 \\
\hline
\multicolumn{5}{l}{*The SRR-5 is merged in SRR-4.}
\end{tabular}
\label{tab5}
\end{center}
\vspace{-0.5em}
\end{table}

\subsection{The Comparison Study with Other Risk Models}\label{s5b}
We implement three other different risk assessment methods, the inverse Time to Collision (1/TTC), the Driving Safety Field (DSF) \cite{wang2015driving}, and the Potential Damage Risk (PODAR) \cite{chen2024quantifying}. They are calculated for each traffic object and constructed as the same data input structure as DSPR. All models also experienced the same training process in Section \ref{s4}. 
Results are shown in Table~\ref{tab6}. DSPR shows the best performance in terms of precision, recall, f1 score, and area under the curve (AUC) score among them, presenting advantages in predicting driver’s SRRs compared to the other three models.

 \begin{table}[t]
\caption{Comparison with other risk assessment models.}
\vspace{-1.5em}
\begin{center}
\begin{tabular}{ccccc}
\hline
\textbf{Model} & \textbf{Precision} & \textbf{Recall} & \textbf{F1-score} & \textbf{AUC}  \\
\hline
\textbf{1/TTC} & 0.5967&0.6522&0.5817&0.8975  \\
\textbf{DSF} & 0.7947&0.7880&0.7779&0.9095 \\
\textbf{PODAR} & 0.7762&0.7677&0.7501&0.9316 \\
\textbf{Ours} & \color{Violet}\textbf{0.8808}&\color{Violet}\textbf{0.8791}&\color{Violet}\textbf{0.8789}&\color{Violet}\textbf{0.9733} \\
\hline
\end{tabular}
\label{tab6}
\end{center}
\vspace{-1.5em}
\end{table}

\subsection{Effectiveness of Semi-supervised Learning Strategy}\label{s5c}
\begin{figure}[t]
\centerline{\includegraphics[width=1\linewidth]{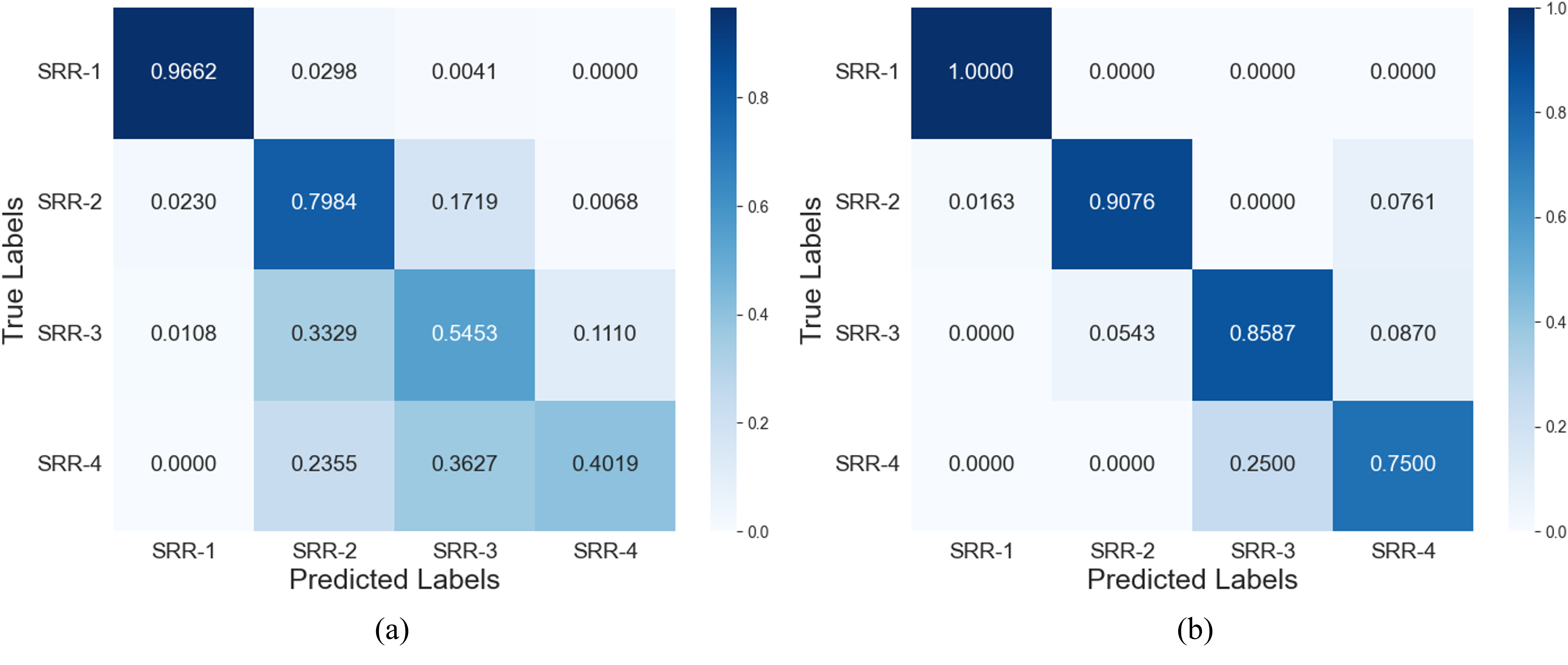}}
\vspace{-0.4em}
\caption{The confusion matrices of predicted results: (a) Results of supervised learning on \(\boldsymbol{R}^{\text{test}}\). (b) Results after label reconstruction and semi-supervised learning strategy on \(\boldsymbol{R}_T^{\text{test}}\).}
\label{fig9}

\end{figure}

We compared the results of supervised training with those of semi-supervised training. As shown in Fig.~\ref{fig9}, the accuracy of the supervised learning on the original test set \(\boldsymbol{R}^{\text{test}}\) is 67.79\%. After the label selection and 5 iterations, the accuracy of the semi-supervised learning on true test set \(\boldsymbol{R}_T^{\text{test}}\) is 87.91\%, improving the accuracy by 20.12\%. Proposed strategy significantly improves the prediction accuracy for SRR-2, SRR-3, and SRR-4. However, there are still some shortcomings in distinguishing between SRR-3 and SRR-4, particularly with SRR-4 often being misclassified as SRR-3. This suggests that there is some disagreement among drivers in SRR-3 and SRR-4, making them difficult to differentiate in certain cases.

\subsection{Comparison studies of Four LSTM Networks}\label{s5d}
We constructed three different networks (Bi-LSTM, CNN-LSTM, TPA-LSTM) to compare with our network. As shown in Table~\ref{tab4}, proposed CNN-Bi-LSTM-TPA overall exhibits better preformance in precision, recall, f1-score and AUC socre. Integrating a CNN network as the preprocessing step before the LSTM network allows for more effective extraction of relationships among input features. The attention mechanism further improves the model's ability to capture temporal dependencies. Combining both techniques can lead to enhanced prediction accuracy.

It is evident that all models achieve the best performance for SRR-1, which has a higher consistency. All models find it challenging to distinguish between SRR-3 and SRR-4 due to driver's subjectivity and randomness in rating, which may lead to varied ratings even for similar scenarios.

\begin{table}[t]
\caption{Comparison studies of Bi-LSTM, CNN-LSTM, TPA-LSTM and our model}
\vspace{-2.5em}
\begin{center}
%\begin{tabular}{|>{\centering\arraybackslash}p{0.62cm}|c|>{\centering\arraybackslash}p{1.08cm}|p{1.28cm}|p{1.08cm}|p{1.08cm}|}
    \resizebox{\linewidth}{!}{
    \begin{tabular}{cccccc}
    \hline
    \multirow{2}*{\textbf{SRRs}} & \multirow{2}*{\textbf{Metrics}} & \multicolumn{4}{c}{\textbf{Model}} \\ 
    \cline{3-6}
    ~ & ~ & Bi-LSTM & CNN-LSTM & TPA-LSTM & Ours \\
    \hline
    \multirow{4}*{} &  Precision & \color{Violet}\textbf{1.0000} & 0.9840 & \color{Violet}\textbf{1.0000} & 0.9840 \\
    SRR & Recall & \color{Violet}\textbf{1.0000} & \color{Violet}\textbf{1.0000} & \color{Violet}\textbf{1.0000} & \color{Violet}\textbf{1.0000} \\
    -1 & F1-score & \color{Violet}\textbf{1.0000} & 0.9919 & \color{Violet}\textbf{1.0000} & \color{Violet}\textbf{1.0000} \\
    ~ & AUC & \color{Violet}\textbf{1.0000} & \color{Violet}\textbf{1.0000} & \color{Violet}\textbf{1.0000} & \color{Violet}\textbf{1.0000} \\
    \hline
    \multirow{4}*{} &  Precision & 0.5833 & 0.5825 & 0.7137 & \color{Violet}\textbf{0.9435} \\
    SRR & Recall & \color{Violet}\textbf{0.9511} & 0.9402&0.9348&0.9076 \\
    -2 & F1-score &0.7231&0.7193&0.8094&\color{Violet}\textbf{0.9252}\\
    ~ & AUC & 0.9800&0.9093&\color{Violet}\textbf{0.9857}&0.9612\\
    \hline
    \multirow{4}*{} &  Precision & 0.5882&0.6054&0.7600&\color{Violet}\textbf{0.7745}\\
    SRR & Recall & 0.5435&0.6087&0.8261&\color{Violet}\textbf{0.8587} \\
    -3 & F1-score &0.5650&0.6070&0.7917&\color{Violet}\textbf{0.8144}\\
    ~ & AUC & 0.8687&0.8816&0.9256&\color{Violet}\textbf{0.9766}\\
    \hline
    \multirow{4}*{} &  Precision & 0.9512&\color{Violet}\textbf{1.0000}&0.9640&0.8214\\
    SRR & Recall & 0.4239&0.3641&0.5815&\color{Violet}\textbf{0.7500}\\
    -4 & F1-score &0.5865&0.5339&0.7254&\color{Violet}\textbf{0.7841}\\
    ~ & AUC & \color{Violet}\textbf{0.9555}&0.8038&0.9205&\color{Violet}\textbf{0.9555}\\
    \hline
    \multicolumn{2}{c}{\textbf{Accuracy}} & 0.7296&0.7283&0.8356&\color{Violet}\textbf{0.8791}\\ 
    \hline
    \end{tabular}
    }
\label{tab4}
\end{center}
\vspace{-2.5em}
\end{table}

\section{Conclusion}\label{s6}
In this paper, we designed a risk-trigger mechanism, spatiotemporal decay, and anisotropy for driver's subjective perceived risk modeling. We conducted a driver-in-the-loop experiment and recruited 20 participants for SRR collection. After selecting high consistency labels among SRRs, we employed a semi-supervised learning strategy with a proposed CNN-Bi-LSTM-TPA network to reconstruct the labels and train the model. Results indicate that our proposed DSPR model presents the highest accuracy in predicting driver's perceived risk compared to the other three risk models. The semi-supervised learning strategy effectively reduces data noise caused by the driver, improving the accuracy by 20.12\%. The proposed CNN-Bi-LSTM-TPA network demonstrates the highest accuracy (87.91\%) compared to other LSTM network structures.

In future research: (1) We will expand the dataset scenarios and participant pool to study driver's perceived risk more comprehensively; (2) We will develop DSPR model tailored to different driving scenarios to enhance the model's ability; (3) We will explore the relationship between driver’s SRRs and DSPR model parameters, with the goal of adaptively optimizing DSPR parameters.

%\begin{thebibliography}{00}

\bibliographystyle{IEEEtran}
\bibliography{IEEEabrv,mylib}

%\end{thebibliography}

\end{document}